\documentclass{article}
\usepackage{aaai19}

\usepackage{times}
\usepackage{xcolor}
\usepackage{soul}
\usepackage[utf8]{inputenc}
\usepackage[small]{caption}

\usepackage{graphicx}
\usepackage{subfig}
\graphicspath{{images/}}
\usepackage{url}
\usepackage{verbatim}

\title{Predicting Hurricane Trajectories using a Recurrent Neural Network}

\author{Sheila Alemany$^1$, Jonathan Beltran$^1$, Adrian Perez$^1$, Sam Ganzfried$^2$\\ 
$^1$School of Computing and Information Sciences, Florida International University\\
$^2$Ganzfried Research\\
\{salem010, jbelt021, apere946\}@fiu.edu, sam@ganzfriedresearch.com}

\begin{document}
\maketitle

\begin{abstract}
Hurricanes are cyclones circulating about a defined center whose closed wind speeds exceed 75 mph originating over tropical and subtropical waters. At landfall, hurricanes can result in severe disasters. The accuracy of predicting their trajectory paths is critical to reduce economic loss and save human lives. Given the complexity and nonlinearity of weather data, a recurrent neural network (RNN) could be beneficial in modeling hurricane behavior. We propose the application of a fully connected RNN to predict the trajectory of hurricanes. We employed the RNN over a fine grid to reduce typical truncation errors. We utilized their latitude, longitude, wind speed, and pressure publicly provided by the National Hurricane Center (NHC) to predict the trajectory of a hurricane at 6-hour intervals. Results show that this proposed technique is competitive to methods currently employed by the NHC and can predict up to approximately 120 hours of hurricane path.
\end{abstract}

\section{Introduction}
\label{se:introduction}
Hurricanes are thermally driven, large-scale cyclones whose wind speeds exceed 75 miles per hour and circulate about a well-defined center~\cite{hurricanes}. They arise from the warm waters of the Atlantic Ocean and the Caribbean Sea and typically travel North, Northwest, or Northeast from their point of origin. Hurricanes' nature of having strong winds, heavy precipitation, and dangerous tides typically result in severe economic disasters and loss of lives. In 1965 Hurricane Betsy caused \$1.5 billion of property damage at Florida and the Gulf States~\cite{hurricanes}. In 2005, Hurricane Katrina resulted in 853 recorded fatalities also in regions of Florida~\cite{katrina}. The prediction of hurricane trajectories allows for civilians to properly evacuate and prepare for these dangerous and destructive storms. Hurricane track forecasting is not simple, however, as hurricanes are highly erratic in their movements.

The development of current seasonal hurricane forecast methods has advanced over the past decades. However, most of the current hurricane trajectory forecast methods are statistical in nature~\cite{statistical-dynamical}. These statistical approaches are limiting due to the complexity and nonlinearity of atmospheric systems. Recurrent neural networks (RNNs) have been recently used to forecast increasingly complicated systems. RNNs are a class of artificial neural networks where the modification of weights allows the model to learn intricate dynamic temporal behaviors. A RNN with the capability of efficiently modeling complex nonlinear temporal relationships of a hurricane could increase the accuracy of predicting future hurricane path forecasts. Development of such an approach is the focus of this paper.

While others have used RNNs in the forecasting of weather data, to our knowledge this is the first fully connected recurrent neural networks employed using a grid model for hurricane trajectory forecasts. The proposed method can more accurately predict trajectories of hurricanes compared to traditional forecast methods employed by the National Hurricane Center (NHC) of the National Oceanic and Atmospheric Administration (NOAA). This paper summarizes our present state of model development. After reviewing the related work and background of recurrent neural networks, we describe in detail the elements of our proposed approach. Subsequently, we compare our results with other hurricane forecasting techniques, including ones employed by the NHC.

\section{Related Work}
\label{se:related}
Scientists are interested in improving the capability of predictive models for tracking hurricanes for the safety of individuals. There exist various forecast prediction models for the tracking of hurricanes. The present-day models vary immensely in structure and complexity. The National Hurricane Center (NHC) of the National Oceanic and Atmospheric Administration (NOAA) uses four main types of models in their path predictions: dynamical, statistical, statistical-dynamical, and ensemble or consensus models.\footnote{NHC Track and Intensity models provided at: \url{https://www.nhc.noaa.gov/modelsummary.shtml}}

Dynamical models, or numerical models, are complex as they require the highest computational power to process physical equations of motion in the atmosphere. Kurihara et al. describe a dynamical model called the Geophysical Fluid Dynamics Laboratory (GFDL) Hurricane Prediction System that was created to simulate hurricanes~\cite{gfdl}. This model created a multiply nested movable mesh system to generate the interior structure of a hurricane and used cumulus parameterization to successfully generate a real hurricane. The GFDL model can be divided into four phases: (1) establish a global model forecast onto grid points of the hurricane model, (2) initialize model through the method of vortex replacement, (3) execute model to create a 72-hour prediction, and (4) provide time series information containing storm location, minimum pressure, maximum wind distribution, a map of the storm track and the time sequences of various meteorological fields~\cite{gfdl}.

Statistical models are light-weight models which only use statistical formulas to discover storm behavior relationships from historical data. The relationships utilized to predict hurricane trajectories are based on many storm-specific features collected, such as location and date of the hurricane. In~\cite{statistical-modelling}, the statistical non-parametric model derives simulations of possible trajectory paths by spatially averaging historical data. This method strives to avoid over-fitting with the maximal amount of historical data by using out-of-sample validation to optimize data averaging. Dynamical systems combined with the statistical relationships allow for models to employ large-scale variables as a set of predictors for hurricane forecast schemes. In~\cite{statistical-dynamical}, they utilized a statistical-dynamical model to predict the trajectory paths of hurricanes in the Atlantic based on the relationship between variability of hurricane trajectories, sea surface temperatures, and vertical wind shear. 

Ensemble or consensus models are a combination of forecasts from different models, different physical parameters, or varying model initial conditions. These models have shown to be more accurate than the predictions from their individual model components on average~\cite{ensemble}. However, there are still many challenges associated with modeling nonlinear spatiotemporal systems and as a result, many of the statistical developments have been left behind by disciplines such as machine learning. In~\cite{neural-oscillatory}, data mining techniques are used to predict the features of hurricanes as it provides a time-series analysis. However, this feed-forward attempt of applying machine learning to nonlinear spatiotemporal processes does not capture time-sequential dynamical interactions between variables of natural events~\cite{bayesian-rnn}. 

In~\cite{sparse-recurrent}, sparse Recurrent Neural Network with a flexible topology is used where the weight connections are optimized using a Genetic Algorithm (GA). It accumulates the historical information about dynamics of the system and uses at the time of prediction. Dynamic Time Warping (DTW) is used to make all the hurricanes uniform, allowing the Recurrent Neural Network to equally learn from each hurricane. However, the use of DTW in this model does not allow for hurricanes whose path is not monotonic. In other words, hurricanes that turn back on itself are not considered. Due to the stochastic nature of hurricane trajectories, Recurrent Neural Networks could benefit from learning from all hurricane paths. Therefore, the accuracy of hurricane trajectory forecasts could benefit from a Recurrent Neural Network that could model the temporal behaviors of a due to the complexity and nonlinearity of the atmospheric systems. 

Due to the complexity and non-linearity of the atmospheric systems and lack of using available hurricane observations, as in sparse Recurrent Neural Network, there is a need for networks with the capability of modeling the temporal behaviors of a hurricane. With the high complexity and continuously increasing the quantity of data collected, linear models are limiting. A recurrent neural network with the capability of modeling any complex nonlinear temporal behaviors of a hurricane could increase the accuracy of predicting future hurricane trajectories.

\section{Recurrent Neural Networks}
\label{se:rnn}
Recurrent Neural Networks (RNNs) are nonlinear dynamical models commonly used in machine learning to represent complex dynamical or sequential relationships between variables~\cite{bayesian-rnn}. Dynamical spatiotemporal processes, such as forecasting hurricane trajectories, represent a class of complex systems that can potentially benefit from RNNs. Similar dynamical spatiotemporal processes have benefited greatly from the application of RNNs~\cite{sample-app1,sample-app2}.

\begin{figure}[!htb]
  \centering
  \captionsetup{justification=centering}
	\includegraphics[scale=0.6]{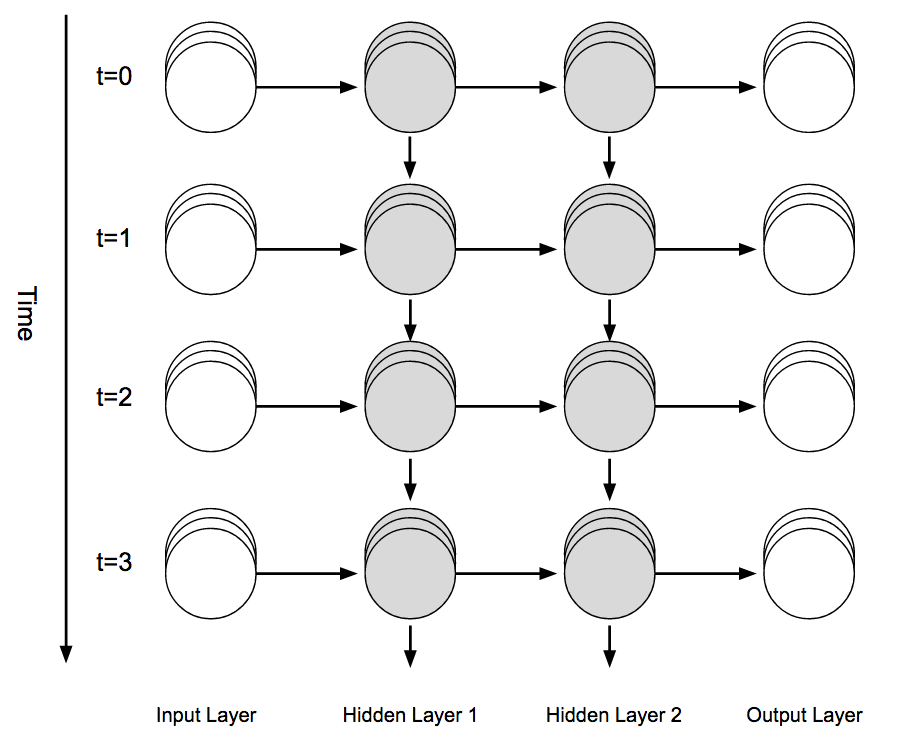}
  \caption{Sample Simple Recurrent Neural Network Architecture}
	\label{fi:RNN-example}
\end{figure}

RNNs are generally fully connected networks, where connection weights are the training parameters. A simple architecture of a deep recurrent neural network arranges hidden state vectors $h_{t}^{l}$ in a two-dimensional grid, where $t=1,\ldots,T$ is the total time the neural network will learn and $l=1,...,L$ is the depth of the network. Figure~\ref{fi:RNN-example} shows a simple RNN architecture with a depth of two.

\begin{figure*}[!ht]
  \centering
  \captionsetup{justification=centering}
  \subfloat[Largest Distance Traveled]{\includegraphics[scale=0.3]{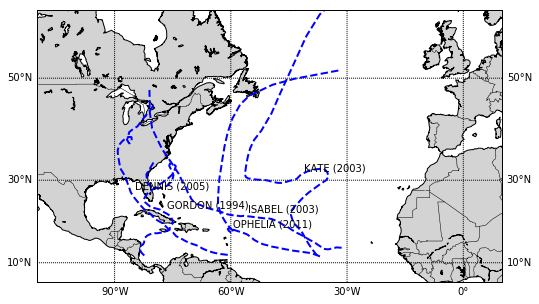}}
  \hfill
  \subfloat[Smallest Distance Traveled]{\includegraphics[scale=0.3]{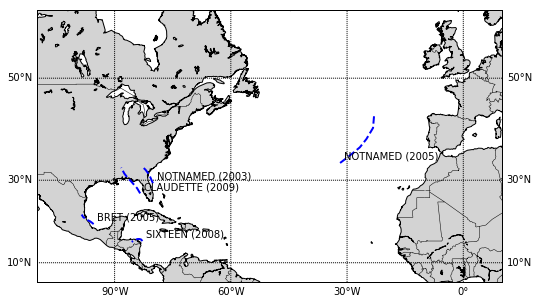}}
  \hfill
  \subfloat[Randomly Selected]{\includegraphics[scale=0.3]{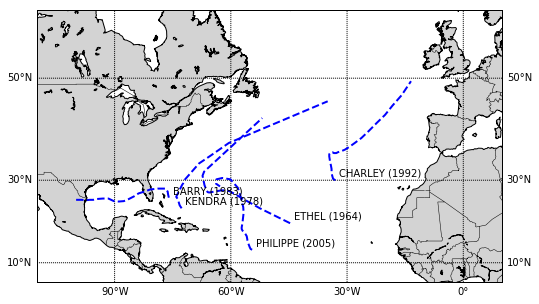}}
  \caption{15 Atlantic Hurricane Trajectories Collected by Unisys Weather$^2$}
	\label{fi:trajectories}
\end{figure*}

The $n_y$-dimensional vector output $Y_{t}$ corresponding to the original $n_x$-dimensional vector input $X_{t}$ is given by:
$$ 
Y_t = g(V^{l} * h_t^{L}) \eqno{(1)}
$$
where $h_t^{L}$ is the final $n_h$-dimensional vector of hidden state variables that is used to predict the output vector, $V^{l}$ is the $n_y \times n_h$ weight matrix, and the function $g(\cdot)$ is an activation function that creates the mapping between the output and the hidden states. The hidden state vectors $h_{t}^{l}$ are defined by:
$$
h_{t}^{l} = f(W^{l} * h_{t-1}^{l} + U^{l} * X_{t}) \eqno{(2)}
$$
\noindent where $W^{l}$ is a $n_h \times n_h$ weight matrix, $U^{l}$ is a $n_h \times n_x$ weight matrix, and the function $f(\cdot)$ is the activation function for the hidden layers~\cite{visualizing-rnn}. The activation function $f(\cdot)$ construes the output and is one of the core components of the neural network architecture~\cite{tong2010genetic}. The nonlinearity of the RNN model is influenced by these activation functions. The weight matrix $W$ models underlying dynamic connections between the various hidden states. Thus, latent nonlinear interactions can effectively be modeled within this framework through $W$. Hidden states extract and provide notable hidden dynamic features from the data and allow the $V$ parameters to appropriately weight these patterns~\cite{bayesian-rnn}. Making recurrent neural networks appropriate and effective at modeling and forecasting the complex atmospheric systems of hurricanes.

\section{Unisys Weather Atlantic Hurricane Data}
\label{se:data}
The raw Atlantic hurricane/tropical storm data used in the study were extracted from the NOAA database.\footnote{Atlantic hurricane tracking by year provided at: \url{http://weather.unisys.com/hurricane/atlantic/}} The data contains all hurricanes and tropical storms from 1920 to 2012. It includes the 6-hourly center locations (latitude and longitude in tenths of degrees), as well as the wind speeds (in knots where one knot is equal to 1.15 mph) and minimal central pressure (in millibars) for the life of the historical track of each cyclone. Tropical storms are cyclones whose wind speeds do not exceed 75mph. These were considered in our model as the behavior of tropical storms are the same as those of fully-developed hurricanes and provides insight on trajectory behavior. The trajectories of several of the storms from the database are depicted in Figure~\ref{fi:trajectories}. We utilized the hurricane points with valid longitude, latitude, wind speed, and pressure values. The 50th percentile of hurricanes contained 21 data points collected in its lifespan. Since a tuple was collected 6-hourly, RNN learning was performed mostly on hurricanes with duration of 126 hours. 

The distance traveled and angle of travel, or direction, was extracted from the latitude and longitude values of each point collected 6-hourly by each hurricane. These augmented parameters allow the neural network to learn about relative rather than absolute parameters. Relative variables provide a measure of relation and representational learning for unseen paths. From the set of hurricanes we are utilizing, the largest hurricane, Hurricane Kate (2003), traveled approximately 6394.7 miles while the smallest hurricane, Hurricane Edouard (1984), traveled approximately 86.5 miles.

\begin{figure}[!htb]
  \centering
  \captionsetup{justification=centering}
  \includegraphics[scale=0.6]{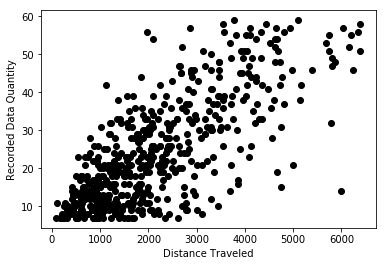}
  \caption{Recorded Data Quantity vs. Distance Traveled}
  \label{fi:data-vs-distance}
\end{figure}

Although not explicitly stated by the original dataset, we can conclude a high correlation between the number of data points collected and the total distance traveled per historical track by the calculated Pearson correlation coefficient of 0.739. This correlation is also evident from Figure~\ref{fi:data-vs-distance}. From this, we can conclude that hurricanes are continuously traveling and therefore, direction and angle of travel continuously provide the recurrent neural network with information about its trajectory behavior.

The direction and angle of travel features were tested against a normal distribution using the Anderson-Darling test. Both features returned a statistic larger than the critical value at a significance level of 0.05. Therefore, the null hypothesis that the features come from a normal distribution is rejected. Using direction and angle of travel as influencing features in our model, the values were normalized to ensure the values of each hurricane were centered around a normal distribution with invariant mean and variances. It has been shown that RNN with normalized data learns from every input equally, generalizing better and converging significantly faster~\cite{cooijmans2016recurrent}. It does so as RNN returns more confident results when learning within the interval [0,1] while being more conservative in its sample. Normalizing is completed using $Z=\frac{X-\mu}{\sigma}$ where $X$ is the original value, $Z$ is the normalized value, $\mu$ and $\sigma$ are the mean and standard deviation of all the values collected by a historical hurricane path, respectively. A normalized value represents the probability that the value could appear in the given historical data, therefore the RNN can modify the weight vectors at each state easily shown by a runtime speedup of the RNN training~\cite{data-mining,cooijmans2016recurrent}. 

\section{Model and Implementation}
\label{se:model}

In this paper, hurricane forecasting is performed by a fully-connected RNN employed over a grid system. The proposed network has the capability to accumulate the historical information about the nonlinear dynamics of the atmospheric system by updating the weight matrices appropriately. This capability makes the RNN suitable for modeling the complex system of hurricane behavior with unobservable states. 

\subsection{Grid Model}
\label{se:grid}
We trained our neural network to learn about a grid model, meaning that the RNN will learn the behavior of a hurricane trajectory moving from one grid location to another. Typical numerical methods contain truncation errors due to computational limitations. As numerical models attempt to increase accuracy and reduce the scale, truncation errors become more excessive~\cite{grid-model}. If minor truncation errors propagate throughout the prediction, this could represent hundreds of miles in potential error. Employing the grid system and reducing the number of possible truncation errors contributes to the improved accuracy of our model by allowing us to control the amount of loss used by prediction.

\begin{figure}[!htb]
  \centering
  \captionsetup{justification=centering}
  \includegraphics[scale=0.42]{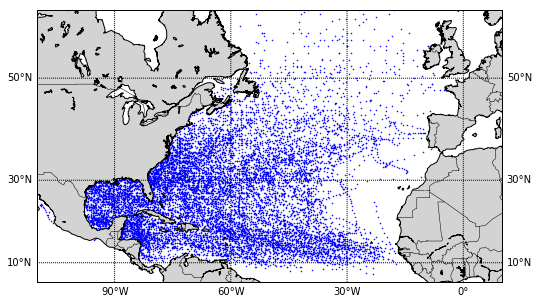}
  \caption{Atlantic Hurricane Points Collected by Unisys Weather}
  \label{fi:hurricane-points}
\end{figure}

Figure~\ref{fi:hurricane-points} shows the recorded latitude and longitude points collected by Unisys Weather Data used to train and test our model. This figure allows us to see the habitual behavior of hurricane trajectories given certain atmospheric information. A refined grid was placed over the latitude and longitude points to reduce truncation errors while allowing for a larger scale model that encapsulates the small-scale features more accurately~\cite{grid-model}. This being an ideal arrangement for RNNs to optimally grasp the complexity of hurricane trajectories.

\subsection{Main RNN Hyperparameters}
\label{se:hyperparameters}
Hyperparameters are preset values optimally selected by the programmers before training begins. The activated values of hyperparameters vary greatly depending on the model being employed, the quantity and quality of the data being used, and the complexity of the model. These hyperparameters are fundamental in encapsulating the nonlinearly and complexity behind forecasting hurricane paths as they influence the decision making steps when updating the weight matrices.

\subsubsection{Grid Boundaries} 
\label{se:boundaries}
The number of grid blocks in our model could be tuned depending on the amount of hurricane data available. The number of grid blocks optimal for the RNN is directly proportional to the number of data points available to train on. Given that we utilized 13,131 total valid data points and 539 hurricane/tropical storm trajectories, we utilized a total of 7,256 grid blocks. The grid blocks were of size 1x1 degrees latitude by longitude. Due to the spherical nature of Earth, the area of each grid block is not uniform in square miles but since most points are centered around Earth's equator, the difference in size for each grid block is negligible.


\subsubsection{Dropout}
\label{se:dropout}

A regularization hyperparameter, known as the dropout value, randomly ignores a percentage of the input to prevent the model from co-adapting to the training set, or overfitting, of hurricane trajectories~\cite{dropout}. This value was set to 0.1 and was tuned and selected using cross-validation, meaning 10\% of each input was ignored when training our RNN.


\subsubsection{Long Short-Term Memory Cell}
\label{se:lstm}
Long Short-Term Memory Cells (LSTM) are a building unit for layers in RNNs. Inside an LSTM, there exist three interacting activation layers, each containing their own individual training parameters. The main purpose of these cells is to remember values over arbitrary time intervals by preventing vanishing and exploding weights throughout the RNN. As a result, LSTMs have shown to provide a significant improvement in RNN performance when applications require long-range reasoning~\cite{visualizing-rnn}. There exists other variants of LSTM for RNNs with the same intentions; however, LSTM was the most successful in storing and retrieving information over long periods of time~\cite{visualizing-rnn}. As hurricane trajectories can span over hundreds of hours, these long-term cells in the hidden layers positively contribute to the RNN's ability to learn about each hurricane. 

\subsubsection{Hidden State Vectors}
\label{se:hidden-state-vectors}

Hidden state vectors, often referred to as hidden layers, isolate notable hidden dynamic features from the input data. The number of hidden layers in RNNs contribute to the complexity of the model. It has been shown that having at least two hidden state vectors returns satisfactory results, but more than three hidden state vectors do not provide significant improvement: increasing the hidden layers, given there are more than three, tends to overfit ~\cite{visualizing-rnn}. Therefore, we employed three hidden layers each with a long short-term memory cell to properly encapsulate the complexity of hurricane trajectory behavior while not overfitting.


\subsection{Network Architecture and Implementation}
\label{se:implementation}

For the network architecture, five total layers were employed. An input layer, three hidden layers as previous described, and an output layer. The input layer takes in a data tuple. A data tuple is a sequence of features containing the wind speed, latitude and latitude coordinates, direction (or angle of travel), distance, and grid identification number. The output layer contains an LSTM building unit along with the dropout value, a dense layer, and the final activation layer. Our model utilized the activation function hyperbolic tangent rather than the frequently used sigmoid and rectified linear unit functions in the output layer. This is so as this activation function allowed the model to output values between [-1, 1] which better models movement in all directions. The output shape of each layer contains the input feature count, the time step, and the output size in that order.

For the implementation of our model, we utilized Keras. Keras is an API that integrates with lower-level deep learning languages such as TensorFlow. It has been increasingly used in the industry and research community with over 200,000 individual users as of November 2017 and large scientific organizations including CERN and NASA\footnote{Keras API: \url{https://keras.io/layers/recurrent/}}. This API provided a sequential recurrent neural network model with input, hidden, and output layers created with the necessary parameters to facilitate the process of creating the RNN. Keras provides a default learning rate hyperparameter of the value 0.001. The learning rate is the rate at which the RNN updates the weights at each hidden layer and is modified using Stochastic Gradient Descent~\cite{empirical}. The learning rate is used to minimize the error and stabilize the process of updating the weight matrices. The model was trained on an NVIDIA GeForce GTX 1060 with 6GB of RAM which allowed the model to complete training in 200 seconds.

\begin{figure*}[ht]
  \centering
  \captionsetup{justification=centering}

  \subfloat[Hurricane ALEX (1998)]{\includegraphics[scale=0.40]{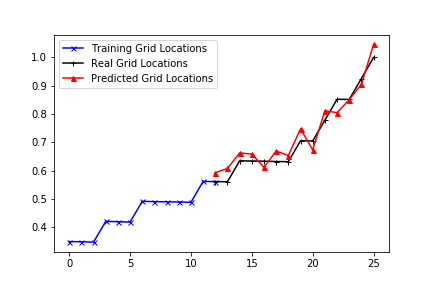}}
  \subfloat[Hurricane DELTA (2005)]{\includegraphics[scale=0.40]{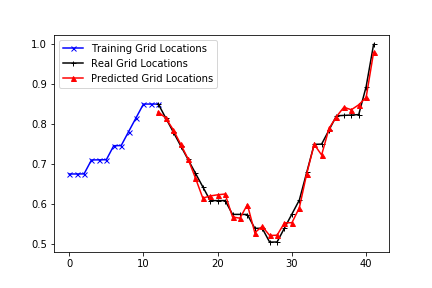}}
  \subfloat[Hurricane SANDY (2012)]{\includegraphics[scale=0.40]{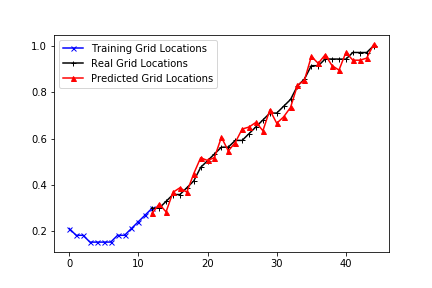}}
  \vspace{-1pt}
  \subfloat[Hurricane HORTENSE (1996)]{\includegraphics[scale=0.40]{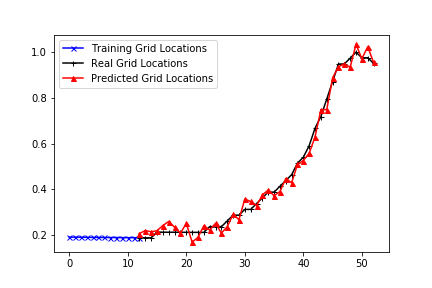}}
  \subfloat[Hurricane IVAN (1960)]{\includegraphics[scale=0.40]{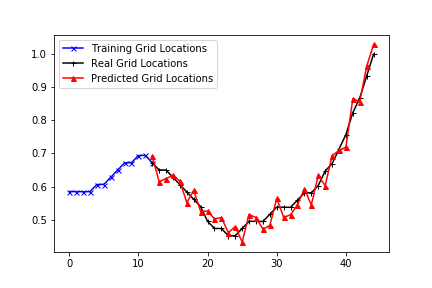}}
  \subfloat[Hurricane PALOMA (2008)]{\includegraphics[scale=0.40]{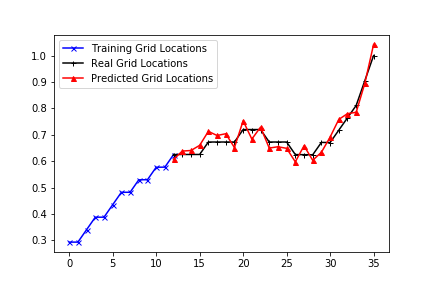}}
  \caption{Randomly Selected Atlantic Hurricane Trajectory Predictions}
	\label{fi:random-predictions}
\end{figure*}


\section{Forecast Results}
\label{se:forecast}
The accuracy of our RNN employed over a grid system is shown for the trajectory predictions of the powerful hurricanes such as ALEX, DELTA, SANDY, HORTENSE, IVAN, and PALOMA. Figure~\ref{fi:random-predictions} shows the forecasted trajectory grid locations of these randomly selected hurricanes against the recorded grid locations from the information provided by NOAA's Unisys Atlantic Hurricane Weather Data. The random selection of these hurricanes occurred on the largest 50th percentile of hurricanes in the testing set. This was selected as so because hurricanes that are longer in duration and distance are better to show the model's prediction capabilities than with shorter hurricanes. Of these selected hurricanes, Hurricane SANDY was the deadliest, most destructive, and costliest hurricane. Due to its excessive wind field, large storm surge, and unusual track into the population-dense area, the damage was about \$75 billion~\cite{sandy}. Thousands of business were damaged or forced to close in addition to the 650,000 homes that were damaged or destroyed. None of the current methods employed by the NHC predicted it would reach the Northern-most part of the U.S and affect New York the way it did until it was too late for most to evacuate. At least 159 people were killed along the path of the storm.

The data provided by Unisys Weather data was divided where 85\% of the total hurricanes were used for training, and 15\% were used for testing the accuracy of our model. In other words, the training set and testing contained 27,477 and 4,850 individual data tuples, respectively. Each data tuple containing a sequence of features of size 5 comprised of wind speed, latitude and longitude coordinates, direction, and grid identification number. When the RNN is deriving a model to predict nonlinear and complex systems, predictive quantification and validity is essential when testing on a dataset different from the training dataset~\cite{validation-importance}. As a result, validation of the training set was completed on 10\% of the 85\% training set, or 2,747 data tuples. At the time of testing, hurricanes were fed into the RNN one hurricane, or tropical storm, at a time. Figure~\ref{fi:random-predictions} shows some examples that the RNN was able to successfully encapsulate, model, and forecast the future trajectory paths of hurricanes at 6-hour intervals. Figure \ref{fi:random-predictions} shows that the grid locations predicted followed the similar trajectory behavior as the real hurricane trajectory.

As was stated in the Related Work section, the sparse Recurrent Neural Network with a flexible topology was trained and tested using the Unisys Weather dataset. This method, although presented impressive results, lacks in modeling and forecasting hurricane behaviors that occur frequently in nature. Due to their use of dynamic time warping (DTW), they are unable to train or test on hurricanes that contain loops. In Table~\ref{ta:MAE}, we compare the Mean Absolute Error (MAE) results they presented with the MAE results of scaled values we predicted. The methods referred to in the related work are not open source and are difficult to reproduce. Therefore, we compared with~\cite{sparse-recurrent} as they used the same dataset and tested with similar hurricanes. However, their predictions were in the form of latitude and longitude with a calculated MAE for each. Our method which employed the grid system returns grid locations as converting back to trajectory paths reduces accuracy. The MAE calculated between the scaled predicted grid block value and the real grid block value. Our grid boundaries are located at a 1x1-degree scale to latitude and longitude. Although we are comparing errors with respect to grid block values for our approach versus errors over the latitude and longitude values for the sparse RNN approach, these two methods can be compared from the provided MAE values. The most appropriate comparison from the table is between the Grid-Based RNN MAE and the average of latitude and longitude MAE for sparse RNN. We can clearly see that our method properly encapsulated the nonlinearity and complexity of hurricane trajectories.

From Table \ref{ta:MAE}, one can see that we have significantly less mean absolute values than the sparse RNN method. This is because their model is incapable of predicting hurricane trajectories that are monotonic. Due to this, their training set is much smaller, and since only monotonic hurricanes were used for training, it can only predict hurricanes without loops. As a result, their method is an impractical solution to predicting hurricanes trajectories due to hurricane's dynamic tendency.

\begin{table}[ht]
\centering
\renewcommand{\arraystretch}{1.3}
\begin{tabular}{*{4}{c}} \hline \hline
Hurricane &DEAN &SANDY &ISAAC\\ \hline \hline
\textbf{Grid-Based RNN} &\textbf{0.0842} &\textbf{0.0800} &\textbf{0.0592}\\ \hline 
Sparse RNN Latitude &0.8651 &0.2500 &0.7888\\ 
Sparse RNN Longitude &0.0572 &0.5949 &0.3425\\ 
\textbf{Sparse RNN Average} &\textbf{0.46115} &\textbf{0.42245} &\textbf{0.56565}\\ \hline \hline
\end{tabular}
\caption{Mean absolute error using our grid-based approach and the sparse RNN approach. Errors of our approach are with respect to the grid created from the latitude and longitude data while the sparse RNN errors are with respect to the direct latitude and longitude data.}
\label{ta:MAE}
\end{table}

\begin{figure}[!ht]
  \centering
  \captionsetup{justification=centering}
  \includegraphics[scale=0.55]{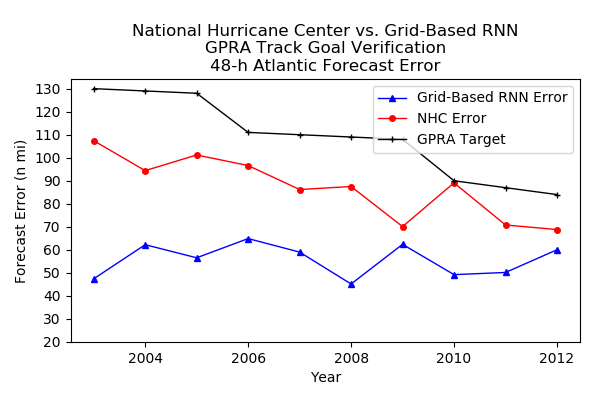}
  \caption{Comparison of Grid-Based RNN against National Hurricane Center 48-hour forecasts. Both methods perform better than required by the Government Performance and Results Act (GPRA), and Grid-Based RNN outperforms currently employed NHC methods.$^4$}
	\label{fi:predictions-all}
\end{figure}

To compare with the methods employed by the National Hurricane Center (NHC), the annual errors collected from all tropical cyclones are plotted against the Government Performance and Results Act (GPRA)\footnote{National Hurricane Center GPRA Track Goal Verification: \url{https://www.nhc.noaa.gov/verification/verify8.shtml}}. These annual errors are calculated for all 48-hour forecasts for the prediction techniques employed by the National Hurricane Center. The track goal is specifically for only 48-hour forecasts, regardless of having forecasts ranging from 6- to 120-hour forecasts, as it is more important for disaster management and citizen preparedness actions. The comparison against GPRA started in year 2000. Figure \ref{fi:predictions-all} shows the performance of the Grid-Based RNN against the NHC annual errors from year 2000 to 2012. As shown in the figure, our Grid-Based RNN performed lower than currently employed NHC techniques due to the fact that the refined grid reduces truncation errors (a common occurrence is statistical-dynamical models utilized by NHC) while allowing for a more extensive scale model that encapsulates the small-scale features more accurately~\cite{grid-model}. As a result, the RNN learns the behavior from one relative location to the next and not general hurricane trajectories which differ highly given the different nonlinear and dynamic features.

The overall regression slope of the error for the Grid-Based RNN is less compared to NHC methods due to the fact that the training set contained hurricanes of years ranging from 1920 to 2012 due to the fact that the dataset was unordered when the training set and testing set were created and remained unordered during the training process of the recurrent neural network. The largest contribution to the forecast error for the Grid-Based RNN is due to the error penalty of 50 km when converting from the 1x1-degree grid area to latitude and longitude coordinates. 

In addition to an improvement in forecast error, statistical-dynamical models used by NHC, require many hours only to make a single prediction utilizing the world’s most advanced supercomputers. The application of neural networks is bleeding edge for hurricane track predictions as it returns accurate forecasts significantly faster (our model completed the training in 200 seconds) in comparison to the hours it takes using the models in practice. The run-time complexity of these algorithms is vital for up-to-date forecasts. A trained neural network can make predictions instantly. As the quantity of data increases and the number of parameters increase given new sensing technologies, statistical-dynamical forecasting techniques become impractical.

\subsection{Limitations and Further Enhancement}
\label{se:limitations}

Our model's final results were grid location prediction coordinates (rather than latitude and longitude) with high accuracy, as shown in Figure~\ref{fi:predictions-all}. However, when we extract individual hurricane trajectories from grid locations and converting from grid locations to latitude and longitude coordinates increases the margin of error. This increase in margin of error could be up to 50 km. The size of the grids is not restricted to be a 1x1-degree scale to latitude and longitude. However, the size of the size of the grids is directly correlated to the number of recorded data points per square kilometer. As a result, a more extensive dataset would allow for smaller grid locations which also reduces the error at the time of conversion. Currently, maintaining a 50 km margin of error due to the grid locations being of 1x1-degree scale to latitude and longitude is still competitive with the current methods employed by the NHC as most of the methods used have at least approximately 50 km margin of error~\cite{leonardo2017verification}.

We intended to directly compare the model's capabilities to capture the nonlinearity of hurricane trajectory features to make predictions. In future work, we will explore the application of an artificial neural network to accurately and adequately convert from grid locations to latitude-longitude coordinates to minimize the conversion error. Also, the implementation of a Bayesian neural network in combination with our grid-based RNN could increase accuracy as Bayesian models could quantify the uncertainty of a prediction. This uncertainty parameter is also valuable information in hurricane trajectory predictions. 

\section{Conclusion}
\label{se:conclusion}

We proposed a recurrent neural network employed over a grid system with the intention to encapsulate the nonlinearity and complexity behind forecasting hurricane trajectories and potentially increasing the accuracy compared to operating hurricane track forecasting models. Our model predicts the next hurricane location at 6 hours, as Unisys Weather data collect the hurricane points with this time frequency. The mean-squared error and root-mean-squared error were 0.01 and 0.11, respectively, for both the training and testing set. The main advantage over the proposed method and the previous technique (“sparse RNN”) is that their approach does not work on hurricanes that loop in behavior. All storms used for training and testing are monotonic, or cannot turn back on themselves, an assumption that is not always true in hurricane trajectory behavior. Our grid-based RNN can be trained and predict hurricanes of any type. 

The high accuracy when comparing against currently employed NHC methods in predicting hurricane trajectories is due to the augmented features of direction and distance traveled in combination to the employed grid. This is because a refined grid reduces truncation errors (a common occurrence is statistical-dynamical models) while allowing for a more extensive scale model that encapsulates the small-scale features more accurately~\cite{grid-model}. Also, the RNN learns the behavior from one physical location to the next and not general hurricane trajectories which differ highly given the different nonlinear and dynamic features. As a result, this model is ideal for RNNs to grasp the complexity of hurricane trajectories optimally. 

Although the idea of grid-based models and recurrent neural networks are not new, the combination of both to model complex nonlinear temporal relationships is a novel contribution. Future researchers might benefit from this paper as this grid-based model is suitable for any application that requires predictions on Euclidean spatiotemporal time series data. Also, as the size of data increases, a model capable of quickly processing and accurately predicting hurricane trajectory information is crucial for the safety of individuals. The high complexity of currently employed methods by the NHC takes hours to make a single prediction and will come to be highly impractical. Future work will consist of using deep learning to convert from grid locations to latitude and longitude coordinates to reduce the conversion rate from grid locations to latitude and longitude coordinates regardless of the size of the dataset. Overall, the intended purpose of this paper is to introduce deep learning for practical hurricane forecasting to increase accuracy while being more lightweight than the statistical-dynamical methods currently employed by the NHC.

\clearpage
\bibliographystyle{aaai}
\bibliography{aaai19}

\end{document}